\documentclass[research]{fcs}

\usepackage{amsmath,amsfonts}
\usepackage{algorithmic}
\usepackage{algorithm}
\usepackage{amsmath}
\usepackage{array}
\usepackage{textcomp}
\usepackage{stfloats}
\usepackage{url}
\usepackage{verbatim}
\usepackage{graphicx}
\usepackage{cite}
\usepackage{multirow} 
\usepackage{float}
\usepackage{bm}
\usepackage{amssymb,amsfonts}
\usepackage{nicefrac}     
\usepackage{microtype}      
\usepackage{lipsum}
\usepackage{color}
\usepackage{booktabs}
\usepackage{epsfig}
\usepackage{epstopdf}
\usepackage[normalem]{ulem}
\useunder{\uline}{\ul}{}
\usepackage{hyperref}

\newcommand{\ours}{BooG}

\title{Boosting Cross-Domain and Cross-Task Generalization for Text-Attributed Graphs from Structural Perspective}
\author[1]{Yao CHENG}
\author[1]{Jiapeng ZHU}
\author[1]{Yige ZHAO}
\author[1]{Jianxiang YU}
\author[1]{Jiaqi TAN}
\author*[1]{Xiang LI}
\address[1]{School of Data Science and Engineering, East China Normal University, Shanghai 200062, China}

\fcssetup{
  received       = {month dd, yyyy},
  accepted       = {month dd, yyyy},
  corr-email     = {xiangli@dase.ecnu.edu.cn},
}
\begin{abstract}
Graph models based on large language models (LLMs) have recently garnered considerable attention due to its significant success.
Although existing methods resort to LLMs to learn unified semantic representations across domains, 
they disregard the unique structural characteristics of graphs from different domains.
To address this problem,
in this paper, 
we boost graph models from structural perspective and propose \ours. 
The model constructs virtual super nodes to unify structural characteristics of graph data from different domains. 
Specifically, 
the super nodes fuse the information of anchor nodes and class labels, where each anchor node captures the information of a node or a graph instance to be classified.
Instead of using the raw graph structure,
the super nodes, along with
virtual edges, establish a standardized aggregation mechanism that fuses rich information from neighborhoods and
associated class labels, accommodating graph structural characteristics inherent to different domains.
Additionally, 
we propose a novel pre-training objective based on contrastive learning, 
which learns more expressive representations for graph data and generalizes effectively to different domains and downstream tasks.
Experimental results on various datasets and tasks demonstrate the superior performance of \ours.  
We provide our code and data here: 
\url{https://github.com/cy623/BooG}.
\end{abstract}

\begin{document}

\section{Introduction}
Graph learning has achieved outstanding results in a wide range of application fields in recent years, including social networks~\cite{sankar2021graph,leskovec2010predicting}, citation networks~\cite{lv2021we}, molecular graphs~\cite{dwivedi2023benchmarking,zitnik2017predicting} and recommendation systems~\cite{he2020lightgcn,song2024xgcn}.
For graph data that encompasses intricate structural information,
graph neural networks (GNNs)~\cite{GAT,GCN,GIN,zeng2023aspect,luo2023bgnn,gasteiger2018predict,hamilton2017inductive,abu2019mixhop,he2022block,jin2021universal,gong2024survey} 
employ sophisticated message-passing mechanisms to effectively learn consistent representations for instances that share similar characteristics, thereby enabling the application of these representations to a wide range of downstream tasks~\cite{zhu2023combat, yan2023efficient,xiao2023graph,wu2023dynamic,zhang2024meta,liu2024evolvekg,liu2024uncertain,tang2024semantic,wu2024graph,wang2022powerful,liu2023beyond,pandit2007netprobe,zhu2021graph,ding2021diffmg,oloulade2021graph,zhang2024introducing
},
such as node classification~\cite{GCN}, graph classification~\cite{GIN} and link prediction~\cite{zhang2018link}.

Although general GNN models have achieved widespread success, 
they often face limitations due to training labels and can only perform a single task.
In practical applications, accurate ground truth labels often come at a high cost~\cite{jin2021automated}. 
To address these challenges,
graph self-supervised learning (SSL)~\cite{DGI,graphcl,jin2021automated,hu2020gpt,qiu2020gcc,graphmae} utilizes the structure or characteristics of the data itself for learning, without the need for explicitly providing labels for the training instances.
For example, 
graph contrastive learning (GCL)~\cite{graphcl} supplements supervised signals by comparing the similarity and dissimilarity between different instances in the graph data.
Graph SSL methods typically employ a learning pipeline involving ``pre-training'' and ``fine-tuning''. 
By designing appropriate pre-training tasks,
the model is enabled to generalize to downstream tasks after fine-tuning.

Although graph SSL has been demonstrated to effectively learn representations of graph data during the pre-training phase, 
the disparity between pre-training tasks and downstream tasks can result in suboptimal performance~\cite{liu2023pre}.
With the development of large language models (LLMs), prompt learning has been proposed to narrow the gap between pre-training and downstream tasks. 
Recently, ``prompt'' has been introduced into graph learning~\cite{gong2023prompt,graphprompt,allinone}. 
Such efforts aim to narrow the gap between pre-training and downstream tasks by designing unified pre-training and prompting templates.
However, 
graph prompt learning fails to assist pre-trained models in improving domain generalization capabilities. 
This is because graph data from different domains often contain specific semantic characteristics. 
For instance, 
features of molecular graphs are typically vectors representing indices of nominal features of atoms, while features of citation networks often consist of bag-of-words vectors describing paper titles and abstracts.

To enhance the generalization capability of graph learning for cross-domain graph tasks,
graph models based on LLMs aim to learn unified semantic representations on graph data from different domains by text-attribute graphs (TAGs)~\cite{ofa,opengraph,unigraph}. 
However, 
these approaches are still constrained by the unique structural characteristics of graph data from different domains. 
For example, 
citation networks contain citation relationships between academic papers, often exhibiting high levels of clustering, where papers referencing each other may involve similar topics or domains. 
On the other hand, 
chemical molecular graphs typically exhibit fixed topological structures and physical-chemical properties.
The structural characteristics from different domains make it challenging for models to generalize the learned structural knowledge to other domains.

To address the aforementioned challenges, 
in this paper, 
we propose \ours, a pre-trained graph model designed for
\textbf{Boo}sting cross-domain and cross-task  generalization for text-attributed \textbf{G}raphs.
\emph{\ours\ unifies the attributes and structures of graph data across different domains, thereby enhancing the capabilities of the pre-trained model to downstream tasks and data from various domains}. 
Specifically, 
\ours\ first leverages a pre-trained language model (LM) to encode the attributes of the graph data and class labels into a unified representation space. 
Next,
\ours\ unifies graph structures by introducing sub-graphs and super nodes. Sub-graphs offer a consistent task template across all levels of tasks, consisting of anchor nodes and their neighbor nodes. The super nodes, along with virtual edges, establish a standardized aggregation mechanism that fuses rich information from neighborhoods and associated class labels, accommodating graph structural characteristics inherent to different domains.
Additionally, 
we propose a novel pre-training objective based on contrastive learning, which enables the model to learn more expressive nodes/graphs representations that generalize effectively to different downstream domains and tasks.
Our extensive experiments on seven datasets, covering a range of domains and tasks, show that \ours\ outperforms state-of-the-art competitors in most cases. We also conduct ablation studies to validate the necessity of each component in \ours.

\section{Related Works}
\subsection{Graph Neural Network.}
As an effective technique for handling graph-structured data,
GNNs~\cite{GAT,GCN,GIN,APPNP,GraphSAGE,SGC,xu2018powerful,suresh2021breaking,yang2021diverse,jiang2024incorporating,jiang2024graph,tang2024semantic,wu2025d} can effectively capture complex patterns based on input node features and structural information. 
For instance, 
graph attention networks (GAT)~\cite{GAT} utilize attention mechanisms to learn the importance of neighboring nodes and aggregate their information with learned weights. 
Graph Isomorphism Network (GIN)~\cite{GIN} adopts a unique graph encoding method that emphasizes distinguishing non-isomorphic structures, enhancing the model's representational capacity. 
However, 
the effectiveness of GNNs often relies on a large number of training labels, implying high labeling costs. 
Additionally, 
the design of GNNs is often tailored to specific tasks, 
making it challenging to generalize to other downstream tasks. 
For instance, 
GNNs designed for node classification tasks often focus more on local information of the instances, while methods designed for graph classification tasks require more attention to the global information of the instances.

\subsection{Graph Self-Supervised Learning.}

Graph SSL~\cite{DGI,graphcl,jin2021automated,hu2020gpt,qiu2020gcc,graphmae} is primarily used to address the problem of lacking labels in graph learning tasks. 
These methods typically employ learning pipelines that include ``pre-training'' and ``fine-tuning'' stages. 
Through carefully designed contrastive learning or generative pre-training tasks, the model can generalize the learned knowledge to downstream tasks after fine-tuning.
For example, 
GraphCL~\cite{graphcl} utilizes the pre-training of graph models by applying a self-supervised contrastive learning task on learned node embeddings. 
DGI~\cite{DGI} introduces a self-supervised pre-training task aimed at maximizing the mutual information between local node views and global graph views.
While graph SSL has been demonstrated to effectively learn instance representations during the pre-training phase, 
the disparity between pre-training tasks and downstream tasks results in suboptimal performance.

Inspired by advancements in the field of NLP, graph prompting learning~\cite{gong2023prompt,graphprompt,allinone} has been proposed to bridge the gap between pre-training and downstream tasks.
For example,
GraphPrompt~\cite{graphprompt} introduces a learnable prompt layer that can automatically identify key information in the pre-trained model for downstream tasks.
However, 
graph prompt learning fails to assist pre-trained models in improving domain generalization capabilities. 

\subsection{Graph Foundation
Model based on LLM.}
The advent of foundation models first became a reality in NLP due to the development of LLMs. 
Foundation models are trained on extensive data and can adapt to a wide range of data and downstream tasks~\cite{liu2023towards}. 
Existing graph models~\cite{ofa,unigraph,opengraph,zerog,zhao2024graphany,shi2024graph} that integrate LLMs can generally be divided into two categories:
(1) Methods with LLMs as the backbone perform various graph tasks by converting graphs into text or tokens for LLMs utilization. 
(2) Methods with GNN as the backbone typically use pre-trained LLMs to obtain unified representations for instances, 
enhancing the representation capacity of GNNs and improving generalization. 
For example, 
UniGraph~\cite{unigraph} proposes a cascade architecture of LM and GNN, along with a self-supervised training objective based on Masked Graph Modeling (MGM), 
and introduces graph instruction fine-tuning using LLMs to achieve zero-shot prediction capability. 
OFA~\cite{ofa} utilizes a pre-trained LM to align feature spaces of graphs from different domains, enabling supervised training across all graphs.
However, 
these approaches are still constrained by structural characteristics of graph from different domains.

\section{Preliminaries}

\subsection{Text-Attributed Graphs}
Let $\mathcal{G} = \mathcal{(V, E, T_{V},T_{E})}$ denotes a TAG,
where $\mathcal{V}$ represents the set of nodes and $\mathcal{E}$ represents the set of edges.
On TAGs, there is a sequential-text feature $t_{v}\in \mathcal{T_{V}} $ associated with each node $v\in \mathcal{V}$.
Similarly,
for each edge $e_{vu} \in \mathcal{E}$ connecting nodes $v$ and $u$, 
there is a sequential-text feature $t_{e_{uv}} \in \mathcal{T_{E}}$.
It is possible for a TAG to have only $\mathcal{T_{V}}$.
Let $\mathcal{D} = (\mathcal{S}, \mathcal{L}, \mathcal{T_{L}})$ denote a graph dataset,
where the sample set $\mathcal{S}$ may contain one or more graphs and $\mathcal{L}$ represents the set of classes.
In this paper,
there is a text feature $t_{l} \in \mathcal{T_{L}}$ associated with each class $l \in \mathcal{L}$ on each dataset.

\begin{figure*}[t]
  \centering  \includegraphics[width=1.0\linewidth]{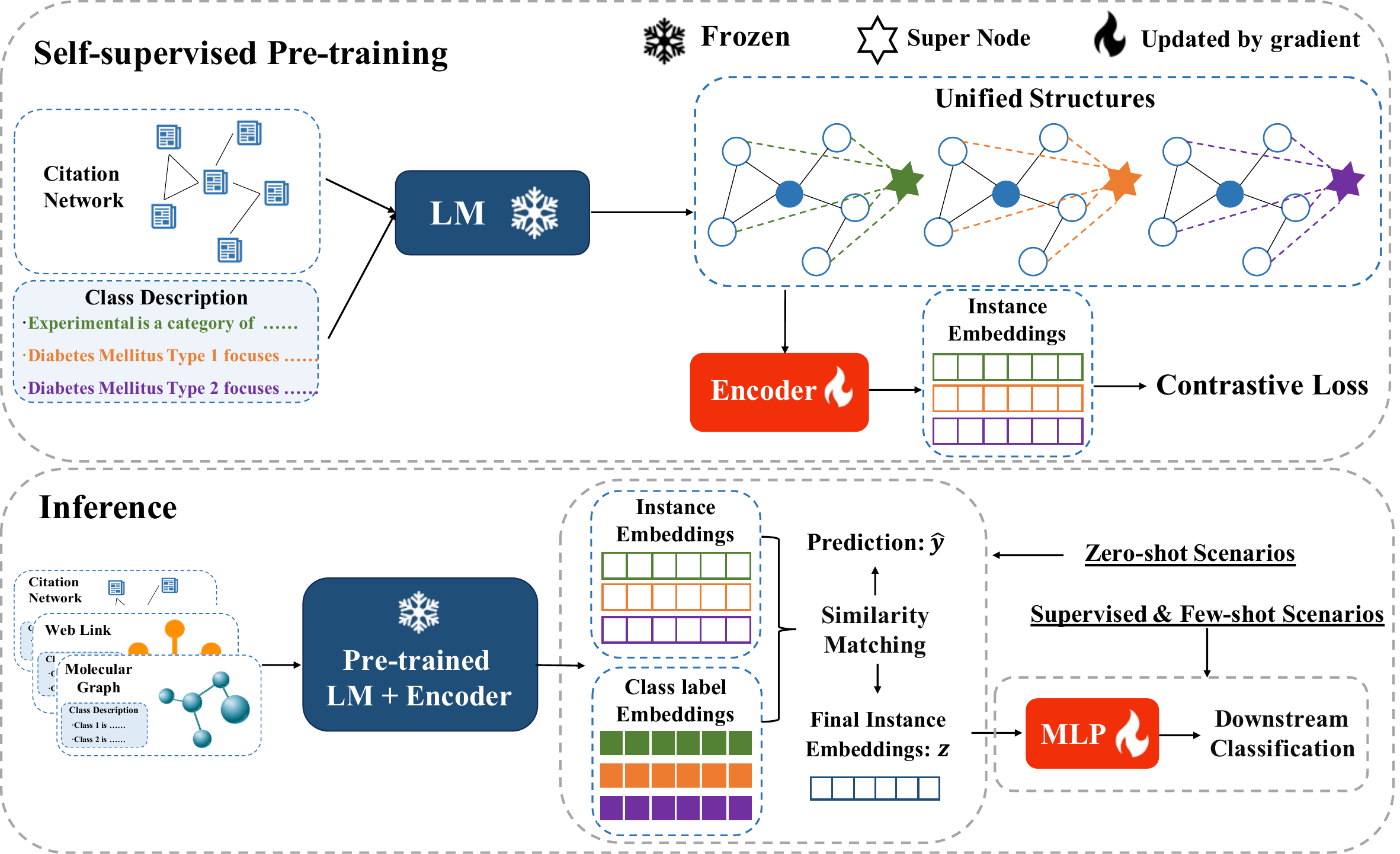}
  \caption{The overall process of \ours.
  The model consists of two parts: A pre-trained model based on self-supervised manner and a downstream classifier implemented with a MLP.
  The model's input includes text attribute graphs and class descriptions. 
  \ours\ first utilizes a pre-trained LM to unify different graph data and standardizes the input for node-level and graph-level tasks as sub-graphs. 
  Subsequently,
  \ours\ introduces super nodes to establish a standardized aggregation mechanism that fuses rich information from neighborhoods and associated class labels.
  We freeze the parameters of the pre-trained model and provide the final instance representations through similarity matching. 
  In particular, the similarity matching process can serve as zero-shot learning to predict unseen instances. 
  For supervised learning and few-shot learning scenarios, 
  \ours\ freezes the parameters of the pre-trained model and generalizes the capabilities of the pre-trained model to multiple downstream tasks by adjusting the parameters of the MLP.
  }
  \label{fg:ours}
\end{figure*}

\subsection{Message Propagation}
Most graph neural networks follow the message-passing mechanism, which consists of two main steps: aggregation and update.
We take an arbitrary node $v_i$ in the $k$-th layer as an example.
The first sub-step is to aggregate information from a node's neighbors,
which is given as:
\begin{equation} 
\hat{h}_{i}^{(k)}=\texttt{AGGREGATE}(h_{j}^{(k-1)}, \forall v_{j}\in \mathcal{N}_{i}). 
\end{equation}
After that,
the second sub-step is to update node embeddings:
\begin{equation} 
h_{i}^{(k)}=\texttt{UPDATE}(h_{i}^{(k-1)}, \hat{h} _{i}^{(k)}). 
\end{equation}
In this paper, 
we implement the encoder by learning information aggregation on the newly constructed structure.

\subsection{Learning Scenarios}
In this paper, 
we propose a cross-domain and cross-task graph model \ours.
This model consists of two parts: A pre-trained model based on self-supervised manner and a downstream classifier implemented with a Multi-Layer Perceptron (MLP).
The pre-trained model, denoted as $f_{\theta }$, operates on a TAG and generates an embedding for each sample (a node or a graph).
We freeze the pre-trained model's parameters when generalizing to downstream tasks.
Our model's generalization ability is evaluated across domains using unseen datasets through three different machine learning problems, and we adopt three graph-related tasks, including node classification, graph classification, and link prediction.
We split the dataset into the train, validation, and test sets, denoted as $\mathcal{D}_{train}$, $\mathcal{D}_{val}$, and $\mathcal{D}_{test}$, respectively.
Their label sets are denoted as
$\mathcal{Y}_{train}$, $\mathcal{Y}_{val}$, and $\mathcal{Y}_{test}$. 
In \textbf{supervised learning} setting,
the classifier will be trained on the training set $\mathcal{D}_{train}$, with the optimal parameters selected based on the validation set $\mathcal{D}_{val}$, and its performance will be finally evaluated on the test set $\mathcal{D}_{test}$.
For \textbf{few-shot learning},
$N$-way $K$-shot tasks evaluate the in-context learning ability to apply the pre-learned knowledge to a new domain and task with $N$ classes, each represented by only $K$ labeled examples.
For \textbf{zero-shot learning}, 
$K$ is set to 0, indicating no prior exposure to support instances from the target classes. 
This setting aims to evaluate a pre-trained model’s ability to generalize and apply its learned knowledge to unseen data categories.

\vspace{-0.2cm}
\section{Methodology}
In this section, we present the \ours\ framework, as illustrated in Figure~\ref{fg:ours}.

\begin{figure*}[h]
  \centering  \includegraphics[width=0.9\linewidth]{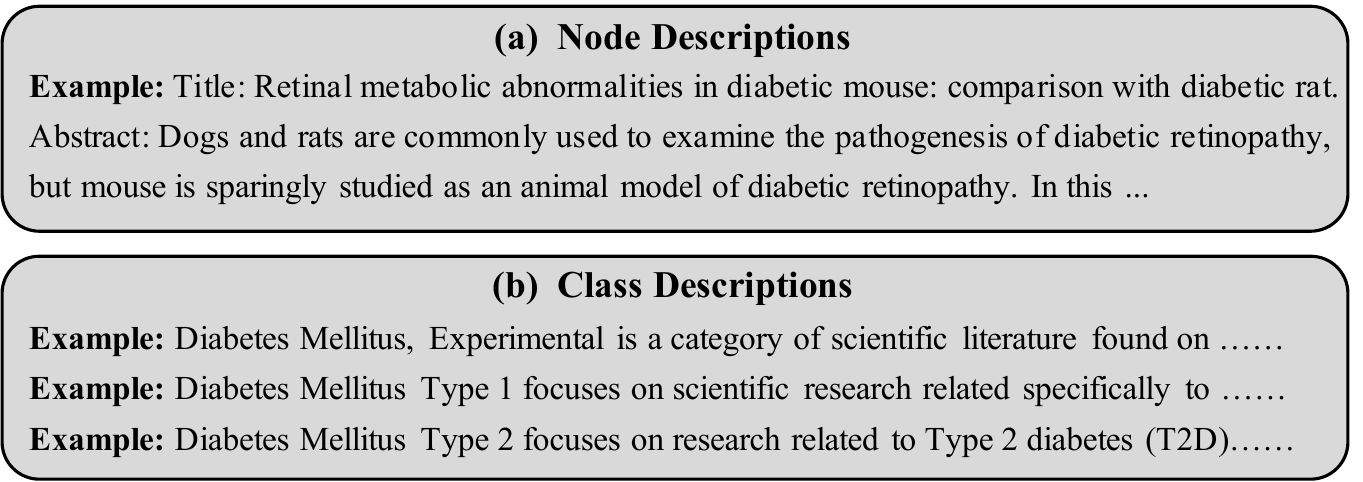}
  \caption{The text format for graph node and class label on Pubmed. 
  }
  \label{fg:text}
\end{figure*}

\subsection{Unifying Graph Attributes}
Most learned GNNs are tailored to specific domains and rely on pre-processed vector features, which limits their ability to generalize across domains due to the inherently distinct and domain-specific nature of training graph features. Fortunately, since graph attributes can be expressed in natural language~\cite{ofa}, in this paper, we unify graph data attributes from different domains by converting input graphs into TAGs and encoding their semantic information using a pre-trained LM. This transformation enables graph nodes to be represented as raw text, as illustrated in Figure~\ref{fg:text}.
Specifically, we employ the Sentence Transformer~\cite{reimers2019sentence} as our pre-trained LM, known for its capability to generate effective sentence embeddings.
Let $t_{v}$ denote the text attributes of a graph node $v$,
then it can be encoded by the pre-trained LM as follows:
\begin{equation} 
    x(v) = \text{LM} (t_{v}) \in \mathbb{R}^{d},
    \label{eq:lm2x}
\end{equation}
where $x(v)$ is the output of the LM, and $d$ is the dimension of the output vector.

In addition, each class $l \in \mathcal{L}$ of each dataset is associated with a text attribute $t_{l} \in \mathcal{T_{L}}$. 
The text attributes $t_{l}$ are also encoded by the same pre-trained LM as follows:
\begin{equation} 
    c(l) = \text{LM} (t_{l}) \in \mathbb{R}^{d}.
    \label{eq:lm2c}
\end{equation}

It is worth noting that \textbf{during \ours's pre-training phase, we rely solely on the textual descriptions of the dataset classes, rather than the class labels of nodes or graphs, thereby establishing a self-supervised learning paradigm}, which will be detailed in the following sections.

\subsection{Unifying Graph Structures}


Graph datasets from different domains exhibit diverse structural characteristics. For example, citation networks often show high clustering coefficients, whereas molecular graphs have fixed, small-world topologies. Moreover, downstream tasks on these datasets can be at different granularities, such as node classification, link prediction, or graph classification. These discrepancies hinder a unified representation learning framework.

\textbf{Sub-graph.}
To accommodate varied domains and tasks under a unified framework, we first convert each input graph into a \textit{sub-graph} centering on an \textit{anchor node}.  
Formally, for a given anchor node $s$, we construct a sub-graph $G_{\text{sub}} = \{s, \mathcal{N}(s)\}$, where $\mathcal{N}(s)$ denotes the neighborhood nodes of $s$.

\begin{itemize}
    \item For \textbf{node/edge-level tasks}, $s$ is a node from the input graph; $\mathcal{N}(s)$ comprises its $k$-hop neighbors.
    \item For \textbf{graph-level tasks}, $s$ represents the whole graph through a pooled embedding $h(s) = R_{\text{graph}}(G)$, and $\mathcal{N}(s)$ contains all nodes in the graph.
\end{itemize}
This step ensures that all tasks are reduced to a common format centered around an anchor node and its context.



\begin{figure*}[ht]
  \centering  \includegraphics[width=1\linewidth]{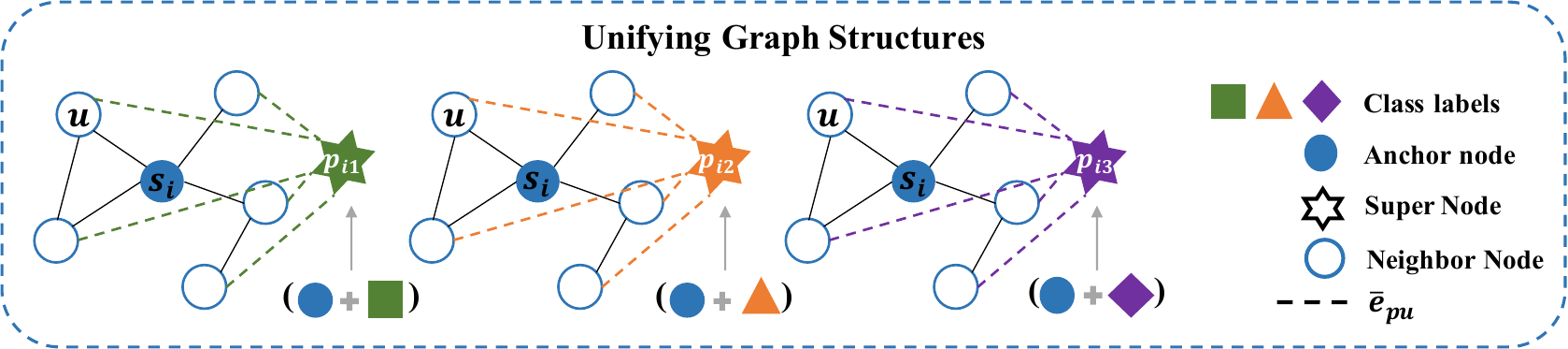}
  \caption{Unifying graph structures by super nodes. 
  }
  \label{fg:super_node}
\end{figure*}

\textbf{Super Nodes.}
We propose \textit{super nodes} to explicitly incorporate \textit{class label information} into the graph structure and to align structure learning across different domains and tasks.

For each anchor node $s_i$ in a sub-graph, we create $C$ super nodes, where $C = |\mathcal{L}|$ is the number of class labels in the dataset. The $j$-th super node $p_{ij}$ is a virtual node representing the \textit{anchor node $s_i$ fused with the class label $l_j$}. The representation is computed as:
\begin{equation}
    h(p_{ij}) = h(s_i) + \alpha \cdot c(l_j),
    \label{eq:target_node}
\end{equation}
where $h(s_i) \in \mathbb{R}^d$ is the embedding of anchor node $s_i$ obtained via the language model, $c(l_j) \in \mathbb{R}^d$ is the embedding of the $j$-th class label, and $\alpha$ is a hyperparameter controlling the weight of class information.
The neighborhood of $p_{ij}$ is defined to be the same as that of $s_i$, i.e., $\mathcal{N}(p_{ij}) = \mathcal{N}(s_i)$. This design allows the super node to aggregate from the same context as the anchor node, while being explicitly tied to a specific class label.

Anchor Node $s_i$ captures instance-specific information.
Super Node $p_{ij}$ hypothesizes ``what if $s_i$ belongs to class $l_j$''.
Multiple Super Nodes provide multiple hypotheses per anchor node, one for each possible class.
These super nodes act as explicit class-guided targets during contrastive learning. Aggregating neighborhood information conditioned on class embeddings facilitates more discriminative and generalizable representations.

\textbf{Unified Structure with Virtual Edges.}
The introduction of super nodes enables a standardized, super node-centered structure that unifies the aggregation process across tasks and domains.
For each anchor node $s$, we insert all of its corresponding super nodes into its neighborhood, as illustrated in Figure~\ref{fg:super_node}.
The neighborhood of each anchor node is obtained through sub-graph sampling to preserve its original structural information.
Super nodes combine class labels with anchor nodes, leveraging these labels to assist anchor nodes in identifying relevant information within their neighborhoods.

To explicitly construct this unified structure, we introduce \textbf{virtual edges} $\bar{e}_{pu}$ connecting each super node $p$ to its neighborhood nodes $u \in \mathcal{N}(p)$.
These virtual edges differ fundamentally from the original edges in the graph $G$ in both purpose and construction.
Original edges reflect the inherent data topology (e.g., citations or chemical bonds), whereas virtual edges are artificially constructed to support standardized aggregation conditioned on class labels.
Specifically, for each super node $p$, $\mathcal{N}(p)$ inherits from the neighborhood of its anchor node $s$, ensuring super nodes access the same neighborhood information.

Original edges propagate node-level information through message passing based on fixed topology.
In contrast, virtual edges allow super nodes to selectively aggregate neighborhood information through learnable attention weights, guiding the prediction towards the correct class label.
This design ensures that the final representations of super nodes are both instance-aware and class-informed.
The virtual edges play a crucial role in BooG’s unified structure design, enabling consistent and comparable aggregation across different datasets and tasks.
While original edges maintain the graph’s inherent topology, virtual edges provide a controlled mechanism for super nodes to gather label-relevant neighborhood information, which is essential for BooG’s generalization capability.

\textbf{Encoder.} 
We then learn to aggregate information within the newly constructed structure using an attention mechanism.
Let $a_{pu}$ denotes the weight of the virtual edge $\bar{e}_{pu}$, indicating
the importance of neighbor node $u$ to super node $p$.
The weights are calculated as follows:
\begin{equation} 
    a_{pu} = \frac{\exp \left( \text{ReLU}(\vec{g}^{\,T} \left [ W_{1} h(p) \parallel W_{2}h(u) \right ]  ) \right) }
    { {\textstyle \sum_{j\in \mathcal{N}(p)}\exp \left(\text{ReLU}(\vec{g}^{\,T} \left [ W_{1} h(p) \parallel W_{2}h(j) \right ]  \right)} },
    \label{eq:att}
\end{equation}
where $h(u)=x(u)$ represents the node features encoded by the LM,
$W_{1}, W_{2} \in \mathbb{R}^{d \times  d}$ and $ \vec{g} \in \mathbb{R}^{2d}$ are learnable parameters,
and $\parallel$ denotes concatenation.
Next, 
we aggregate information for each super node $p_{ij}$ from its neighborhood to obtain
its final representation $\hat{h}(p_{ij})$, computed as:
\begin{equation} 
   \hat{h}(p_{ij}) = W_{3} \left [ \beta h(p_{ij})+ (1-\beta){\textstyle \sum_{u \in \mathcal{N}(p_{ij})}a_{(p_{ij})u}h(u)} \right ] ,
    \label{eq:final_p}
\end{equation}
where $W_{3} \in \mathbb{R}^{d\times d}$ is a learnable parameters and $\beta$ is a hyper-parameter.


\subsection{Self-Supervised Learning Paradigm}
\textbf{Pre-training.}
We pre-train \ours via a contrastive learning objective to enhance its ability to distinguish between different class hypotheses and improve representation generalizability.

For each anchor node $s_i$, we generate $C$ super nodes $\{p_{ij}\}_{j=1}^C$ corresponding to all possible class labels. These super nodes incorporate both the anchor node’s information and individual class semantics through Equation (5). After neighborhood aggregation, we obtain the updated embedding $\hat{h}(p_{ij})$ for each super node $p_{ij}$.

To guide the model toward learning class-discriminative representations, we adopt a normalized temperature-scaled cross-entropy loss~\cite{chen2020simple}, formulated as:
\begin{equation} 
\begin{split}
    & \mathcal{L}_{pre}= \\ 
    & -\frac{1}{C} \sum_{j}^{C} \frac{1}{n} \sum_{i}^{n} log\frac{\exp( \text{sim} (\hat{h}(p_{ij}), h(p_{ij}))/\tau )}
    {\sum_{q=1,q\ne j}^{C} (\exp( \text{sim} (\hat{h}(p_{ij}),\hat{h}(p_{iq}))/\tau))}.
    \label{eq:pre_loss}
\end{split}
\end{equation}
Here, $n$ denotes the number of training instances, $C$ is the number of classes, $\tau$ is a temperature hyper-parameter, and $\text{sim}(\cdot)$ computes cosine similarity.
\textbf{Notably, while the loss function considers the number of classes $C$ in the dataset, it does not necessitate concrete sample-label pairs. 
Our pre-training strategy leverages self-supervised learning via contrastive learning.}

In this framework, the original embedding $h(p_{ij})$ of super node $p_{ij}$—formed by the combination of the anchor node $s_i$ and class label $l_j$—is treated as the \textit{positive sample} for its updated embedding $\hat{h}(p_{ij})$. Meanwhile, embeddings $\hat{h}(p_{iq})$ with $q \neq j$ serve as \textit{negative samples} for the same anchor node $s_i$, as they are associated with different class labels.
This contrastive objective explicitly drives the model to align the aggregated representation $\hat{h}(p_{ij})$ closer to its corresponding class label (positive pair) while pushing it away from other class hypotheses (negative pairs). This alignment forces the model to learn more discriminative features that capture class semantics, thereby enhancing its expressive power and its ability to generalize across tasks and domains.

\textbf{Similarity Matching.}
After pre-training, each anchor node $s_i$ produces $C$ super node embeddings, one for each class label. To select the most appropriate representation for downstream tasks, we apply similarity matching:
\begin{equation}
z_i = \hat{h}(p_{iw}), \quad \text{where} \quad w = \arg\max_{j \in [1, C]} \text{sim}(\hat{h}(p_{ij}), c(l_j)),
\end{equation}
where $c(l_j)$ is the embedding of the $j$-th class label. This step ensures that the final embedding $z_i$ for $s_i$ is selected based on the class label most semantically aligned with the aggregated neighborhood information.





\subsection{Tuning for Downstream Tasks}
\textbf{Zero-shot Scenarios.}
Since the predictions provided by similarity matching do not depend on labels, the pre-trained model can generalize to unseen data, allowing us to treat this process as zero-shot learning.
It is worth noting that
the parameters of the pre-trained model are kept frozen during evaluation under all learning scenarios.
For node/graph classification tasks,
the prediction for the $i$-th node/graph is defined as:
\begin{equation} 
    \hat{y}_{i} = \arg\max_{j}( \text{sim}(\hat{h}(p_{ij}),c(l_{j}))), j\in \left [ 1,C \right ].
    \label{eq:pre}
\end{equation}

For link prediction tasks in the zero-shot setting, we compare the similarity of input node pairs. Pairs with a similarity
exceeding a threshold $T$ are classified as having an edge; otherwise, they are considered not to have an edge.
The link prediction result $\hat{e}_{ij}$ between the pair of nodes $(v_{i},v_{j})$ is computed as follows:
\begin{equation} 
    \hat{e}_{ij} =\begin{cases}
    1, \text{if}~\text{sim}(z_{i}, z_{j})> T
     \\
    0, \text{otherwise}
    \end{cases}.
    \label{eq:pre_link}
\end{equation}

\textbf{Supervised \& Few-shot Scenarios.}
\ours\ uses MLP as the classifier to predict the class of instances. 
Formally, the final prediction for $i$-th input node/graph is given by:
\begin{equation} 
    \hat{y_{i}} =\arg\max(\text{MLP}(z_{i})).
    \label{eq:mlp}
\end{equation}

For link prediction tasks, 
the problem is framed as a binary classification task, where the concatenated representation of node pairs is used as input to the MLP for prediction.
In both supervised and few-shot learning scenarios, we employ cross-entropy loss to train the MLP across all tasks.

\subsection{Time Complexity}
We analyze the time complexity of \ours\ in this section.
For \ours, we denote the maximum sequence length of node textual feature as $L$ and the number of instances as $n$.
For an input graph/node,
the complexity is $O(L^{2}d + Ld^{2})$,
where $d$ is the dimensionality of the embeddings.
The time complexity is $O(n*(L^{2}d + Ld^{2}))$ for processing $n$ instances.
The encoder's time complexity primarily depends on the feature dimensionality and the structure of the graph.
The time complexity of the attention mechanism can be expressed as $O(2d^{2}n +2dn)$.
Next, the complexity of message aggregation is $O(\bar{k}dn+ d^{2}n)$,
where $\bar{k} $ is the average number of neighbors per node. 
The overall time complexity of encoder is $O(k_{1}d^{2}n + k_{2}dn)$, where $k_{1},k_{2}$ are the coefficients.
For similarity matching, 
the time complexity is $O(dnC)$, where $C$ is the number of classes.
Overall, \ours\ has a linear time complexity w.r.t. $n$.

\begin{table*}[ht]
\centering
\normalsize
\caption{Statistics of all text-attributed graph datasets.}
\label{ta:dataset}
\begin{tabular}{ccccccc}
\toprule
\textbf{Dataset}    & \textbf{Domain}   & \textbf{Task}      & \textbf{\#Graphs} & \textbf{Avg.\#Nodes} & \textbf{Avg.\#Edges} & \textbf{Raw Texts}                           \\ \hline
Cora       & Citation & Node/Link & 1        & 2,708       & 10,556      & paper titles and abstracts          \\
Pubmed     & Citation & Node/Link & 1        & 19,717      & 44,338      & paper titles and abstracts          \\
ogbn-arxiv & Citation & Node      & 1        & 169,343     & 1,166,243   & paper titles and abstracts          \\
Wiki-CS    & Web link & Node      & 1        & 11,701      & 216,123     & wikipedia entry names and contents  \\
Ele-Fashion & Product & Node      & 1        & 97,766      & 199,602     & fashion titles \\
PCBA       & Molecule & Graph     & 437,929  & 26.0        & 28.1        & textual descriptions of atoms/bonds \\
HIV        & Molecule & Graph     & 41,127   & 25.5        & 27.5        & textual descriptions of atoms/bonds \\\bottomrule
\end{tabular}
\end{table*}

\subsection{Why \ours\ Works?}

\ours\ boosts model's cross-domain and cross-task generalization ability from two key aspects. First, \ours\ unifies graph attributes by encoding node texts of TAGs and class label descriptions into a shared space using a pre-trained LM, thereby equipping the model to handle graph data from diverse domains. More critically, \ours\ unifies graph structures by introducing the concept of sub-graphs and super nodes. Sub-graphs offer a consistent task template across all levels of tasks, consisting of anchor nodes and their neighbor nodes. The super nodes, along with virtual edges, establish a standardized aggregation mechanism that fuses rich information from neighborhoods and associated class labels, accommodating graph structural characteristics inherent to different domains. Consequently, these designs integrate a supernode-centered graph structure that encapsulates label preferences, making \ours\ compatible with different graph domains and task levels.


Recent studies on graph domain generalization include UniGraph~\cite{unigraph} and OFA~\cite{ofa}.
UniGraph proposes a cascade architecture integrating LM and GNN, utilizing a self-supervised objective based on Masked Graph Modeling. Additionally, it incorporates graph instruction fine-tuning through LLMs to enable zero-shot prediction.
\ours\ freezes LM throughout the process and fundamentally resorts to the unification of graph attributes and structures as the primary vehicle for generalization.
In addition, while OFA leverages an LM to encode node features across domains and introduces the nodes-of-interest (NOI) subgraph and NOI prompt node to unify different graph tasks, 
BooG's super nodes are designed to bind anchor nodes directly with class labels, as expressed in Eq.\eqref{eq:target_node}, which aligns them explicitly within the same latent space, ensuring a more cohesive integration between node features and task-specific labels.

\section{Experiments}

\subsection{Experimental Settings}
\textbf{Datasets.}
We first introduce the datasets used in the experiments. The statistics of these datasets are summarized in Table~\ref{ta:dataset}, and details of each dataset are given below:

\noindent\textbf{\emph{Cora}} is a classic graph dataset widely used in research on GNNs and node classification tasks. 
It consists of 2708 scientific publications classified into one of 7 classes. 
In this dataset, each node represents a scientific paper and each edge represents a citation.
The label of nodes indicates the research field of the paper.
We collect raw text from~\cite{he2023harnessing}.

\noindent\textbf{\emph{Pubmed}} 
is another widely used dataset for graph-based machine learning, particularly in the context of node classification tasks with GNNs.
It consists of 19,717 scientific publications classified into one of 3 classes.
Similar to the Cora dataset, it is a citation network of scientific papers, but it consists of biomedical papers indexed in PubMed, which is a free search engine accessing primarily life sciences and biomedical literature.
We collect raw text from~\cite{he2023harnessing}.

\noindent\textbf{\emph{ogbn-arXiv}}
is part of the Open Graph Benchmark (OGB), a collection of large-scale graph datasets designed to evaluate machine learning algorithms, particularly for graph-based learning tasks.
It is focused on node classification in a citation network of academic papers, specifically from the arXiv repository, which is a preprint server for scientific papers in fields such as physics, mathematics, computer science, and more.
Each node represents a paper, and edges represent the citation relationships between papers.
The task is to predict the 40 subject areas of arXiv CS papers.
We collect raw text from~\cite{hu2020open}.

\noindent\textbf{\emph{Wiki-CS}} 
is another graph dataset in the OGB collection, specifically designed for node classification tasks within the context of social networks.
It is based on Wikipedia articles and represents a graph of scientific papers.
Each node represents a Wikipedia page and each edge represents a reference link.
Each node’s label corresponds to the category of the entry.
We collect raw text from~\cite{ofa}.

\textbf{Ele-Fashion}\cite{zhu2024multimodal} is a node classification dataset derived from the Amazon-Fashion dataset. Each node represents a fashion product, and links indicate user co-purchasing behavior. Textual features are product titles. We collect raw text from\cite{zhu2024multimodal}. The task is to predict product categories, simulating recommendation scenarios.

\noindent\textbf{\emph{PCBA}} is a large-scale chemical bioactivity dataset used for graph machine learning research in cheminformatics and drug discovery. Each node represents an atom of a molecule, and the edges represent the chemical bonds between atoms within the molecule. The primary tasks in the PCBA dataset are to predict the activity of chemical molecules across various bioactivity assays.
We collect raw text from~\cite{ofa}.

\noindent\textbf{\emph{HIV}} is a collection of data used to study issues related to the treatment of HIV (Human Immunodeficiency Virus). It is commonly used in machine learning and data analysis tasks such as drug discovery, drug-virus interaction prediction, and drug resistance prediction.
It is a subset of the PubChem BioAssay dataset.
The task in the HIV dataset involves predicting the activity of chemical molecules against the HIV virus.
We collect raw text from~\cite{ofa}.

For Cora and PubMed, 
we measure the performance of all models on the test sets over 10 random splits as suggested in~\cite{pei2020geom} and report the average accuracy.
For ogbn-arxiv, 
we follow the official splits~\cite{hu2020open}. 
Following the experimental procedure suggested by OGB, we repeat each experiment for 10 times with random seeds and report the average accuracy.
For Wiki-CS, 
we follow the official splits~\cite{mernyei2020wiki} with 20 different training splits and report the average accuracy.
For PCBA and HIV, 
we follow the official splits~\cite{zhao2024gimlet},
repeat each experiment for 10 times with random seeds and report the average accuracy.

\begin{table}[ht]
\centering
\normalsize
\caption{Grid search space.}
\label{ta:hyper}
\begin{tabular}{cc}
\toprule
\textbf{Hyper-parameter} & \textbf{Search space}            \\ \hline
lr                       & \{0.01, 0.02, 0.1, 0.2\}         \\
dropout                  & {[}0.0,0.9{]}                    \\
weight decay             & \{5e-5, 1e-5, 5e-4, 1e-4, 5e-3\} \\
$\alpha$    & {[}0.1,0.9{]}                    \\
$\beta  $   & {[}0.1,0.9{]}                    \\
$\tau  $    & \{0.1,1,10\}                     \\
T                        & {[}0.1,0.9{]}                    \\ \bottomrule
\end{tabular}
\end{table}

\begin{table*}[h!]
\centering
\caption{Experiment results in supervised learning.
We report accuracy (\%) for node/edge tasks and ROC-AUC score (\%) for graph tasks. 
We highlight the best score on each dataset in bold and the runner-up score with underline.
}
\label{ta:supervise}
\resizebox{1\textwidth}{!}{
\begin{tabular}{@{}cc|cccccccccc@{}}
\toprule
\multicolumn{2}{c|}{\textbf{Datasets and tasks}} & \textbf{MLP} & \textbf{GAT} & \textbf{GIN} & \textbf{OFA} & \textbf{GraphCL} & \textbf{DGI} & \textbf{GraphMAE} & \textbf{UniGraph} & \textbf{\ours} \\ \midrule
\multicolumn{1}{c|}{\multirow{5}{*}{Node level}} & Cora & 80.68 & \underline{83.57} & 80.41 & 77.45 & 65.79 & 63.26 & 78.31 & 83.02 & \textbf{83.70}$_{\pm 0.47}$ \\
\multicolumn{1}{c|}{} & Pubmed & 78.62 & \underline{80.46} & 79.26 & 75.16 & 72.30 & 70.81 & 78.24 & 81.57 & \textbf{88.51}$_{\pm 0.31}$ \\
\multicolumn{1}{c|}{} & ogbn-arxiv & 68.31 & 70.24 & 70.55 & \textbf{77.64} & 60.94 & 65.77 & 70.80 & 73.10 & \underline{74.57}$_{\pm 0.61}$ \\
\multicolumn{1}{c|}{} & Wiki-CS & 65.24 & 73.61 & 68.20 & \textbf{78.12} & 58.44 & 60.55 & 64.35 & 72.85 & \underline{75.84}$_{\pm 0.60}$ \\ 
\multicolumn{1}{c|}{} & Ele-Fashion & 58.23 & 62.12 & 65.24 & 68.56 & 55.20 & 60.17 & 64.05 & \underline{70.11} & \textbf{72.34}$_{\pm 0.65}$ \\
\midrule
\multicolumn{1}{c|}{Edge level} & Cora-link & 88.64 & 90.27 & 88.85 & \underline{90.32} & 81.53 & 80.86 & 80.28 & 90.20 & \textbf{93.11}$_{\pm 1.24}$ \\ \midrule
\multicolumn{1}{c|}{\multirow{2}{*}{Graph level}} & PCBA & 52.31 & 50.83 & \textbf{60.03} & 20.89 & 54.30 & 55.17 & 55.67 & 55.18 & \underline{58.26}$_{\pm 1.27}$ \\
\multicolumn{1}{c|}{} & HIV & 60.15 & 62.37 & 70.46 & \underline{71.47} & 65.86 & 62.14 & 67.04 & 71.02 & \textbf{74.50}$_{\pm 1.42}$ \\
\bottomrule
\end{tabular}
}
\end{table*}

\begin{table*}[h!]
\centering
\caption{Experiment results in few-shot learning.
We report accuracy (\%) for node/edge tasks and ROC-AUC score (\%) for graph tasks. 
We highlight the best score on each dataset in bold and the runner-up score with underline.
}
\label{ta:few}
\resizebox{1\textwidth}{!}{
\begin{tabular}{@{}ccc|cccccccccc@{}}
\toprule
\multicolumn{3}{c|}{\textbf{Datasets and tasks}} & \textbf{MLP} & \textbf{GAT} & \textbf{GIN} & \textbf{OFA} & \textbf{GraphCL} & \textbf{DGI} & \textbf{GraphMAE} & \textbf{UniGraph} & \textbf{\ours} \\ \midrule
\multicolumn{1}{c|}{\multirow{10}{*}{Node level}} & \multirow{2}{*}{Cora} & 1-shot & 15.61 & 20.16 & 16.24 & 20.80 & 33.93 & \underline{40.45} & 36.21 & 39.15 & \textbf{44.47}$_{\pm 1.91}$ \\
\multicolumn{1}{c|}{} & & 5-shot & 47.23 & 53.21 & 50.19 & 60.24 & 65.28 & \underline{68.31} & 68.08 & 68.55 & \textbf{70.82}$_{\pm 1.70}$ \\ \cmidrule(l){2-12}
\multicolumn{1}{c|}{} & \multirow{2}{*}{Pubmed} & 1-shot & 15.92 & 18.24 & 19.44 & \underline{30.67} & 22.74 & 20.15 & 24.76 & 29.51 & \textbf{33.85}$_{\pm 1.78}$ \\
\multicolumn{1}{c|}{} & & 5-shot & 45.66 & 50.17 & 48.79 & \textbf{60.27} & 35.16 & 38.61 & 35.65 & 58.24 & \underline{57.56}$_{\pm 1.44}$ \\ \cmidrule(l){2-12}
\multicolumn{1}{c|}{} & \multirow{2}{*}{ogbn-arxiv} & 1-shot & 7.24 & 16.61 & \underline{20.31} & \textbf{22.38} & 18.77 & 16.33 & 17.68 & 19.20 & 20.18$_{\pm 1.74}$ \\
\multicolumn{1}{c|}{} & & 5-shot & 26.86 & 36.37 & 35.65 & \underline{37.69} & 35.81 & 33.45 & 30.27 & 42.90 & \textbf{43.31}$_{\pm 1.53}$ \\ \cmidrule(l){2-12}
\multicolumn{1}{c|}{} & \multirow{2}{*}{Wiki-CS} & 1-shot & 10.70 & 21.55 & 20.33 & \underline{23.18} & 20.55 & 11.25 & 16.84 & 19.77 & \textbf{25.11}$_{\pm 1.07}$ \\
\multicolumn{1}{c|}{} & & 5-shot & 35.15 & 40.86 & 42.15 & \textbf{47.95} & 30.34 & 30.81 & 34.65 & 41.12 & \underline{45.92}$_{\pm 0.85}$ \\ 
\cmidrule(l){2-12}
\multicolumn{1}{c|}{} & \multirow{2}{*}{Ele-Fashion} & 1-shot & 18.50 & 25.62 & 26.40 & 30.22 & 20.15 & 22.84 & 25.95 & \underline{33.10} & \textbf{35.84}$_{\pm 1.21}$ \\
\multicolumn{1}{c|}{} & & 5-shot & 32.64 & 42.01 & 45.32 & 48.51 & 33.77 & 36.44 & 43.13 & \underline{52.45} & \textbf{54.92}$_{\pm 0.78}$ \\
\midrule
\multicolumn{1}{c|}{Edge level} & Cora-link & 1-shot & 30.06 & 50.88 & 52.83 & \textbf{60.27} & 30.57 & 46.11 & 38.82 & 52.31 & \underline{58.27}$_{\pm 1.36}$ \\
\multicolumn{1}{c|}{} & & 5-shot & 63.50 & 76.32 & 77.47 & \underline{80.18} & 55.14 & 61.28 & 75.43 & 79.31 & \textbf{82.41}$_{\pm 1.65}$ \\ \midrule
\multicolumn{1}{c|}{\multirow{4}{*}{Graph level}} & \multirow{2}{*}{PCBA} & 1-shot & 7.55 & 11.04 & 15.24 & \underline{18.22} & 15.11 & 13.18 & 16.31 & 17.02 & \textbf{21.55}$_{\pm 1.18}$ \\
\multicolumn{1}{c|}{} & & 5-shot & 20.16 & 20.37 & \underline{28.46} & 20.14 & 18.87 & 21.63 & 20.62 & 26.55 & \textbf{28.62}$_{\pm 1.54}$ \\ \cmidrule(l){2-12}
\multicolumn{1}{c|}{} & \multirow{2}{*}{HIV} & 1-shot & 30.37 & 33.67 & 40.24 & \underline{45.10} & 36.16 & 35.24 & 33.76 & 41.12 & \textbf{47.74}$_{\pm 1.37}$ \\
\multicolumn{1}{c|}{} & & 5-shot & 53.84 & 50.11 & 57.16 & \underline{63.25} & 46.23 & 50.80 & 38.57 & 63.71 & \textbf{65.48}$_{\pm 1.02}$ \\
\bottomrule
\end{tabular}
}
\end{table*}

\begin{table*}[h!]
\centering
\caption{Experiment results in zero-shot learning.
We report accuracy (\%) for node/edge tasks and ROC-AUC score (\%) for graph tasks. 
We highlight the best score on each dataset in bold and the runner-up score with underline.
Since OFA is first pre-trained on the citation network Cora with training labels, so we did not report the results of OFA on Cora in zero-shot learning.
}
\label{ta:zero}
\resizebox{1\textwidth}{!}{
\begin{tabular}{@{}c|cccccccc@{}}
\toprule
\textbf{Methods} & \textbf{Cora} & \textbf{Pubmed} & \textbf{ogbn-arxiv} & \textbf{Wiki-CS} & \textbf{Ele-Fashion} & \textbf{Cora-link} & \textbf{PCBA} & \textbf{HIV} \\ \midrule
GraphCL & \underline{40.11} & 30.25 & 25.05 & 22.37 & 26.45 & 37.86 & 20.16 & 28.45 \\
DGI & 38.36 & 28.46 & 24.11 & 23.76 & 28.16 & 35.85 & 21.02 & 25.11 \\
GraphMAE & \underline{44.93} & 33.57 & 23.81 & 20.44 & 29.42 & \underline{45.63} & 18.67 & 22.35 \\
OFA & - & \underline{35.07} & \underline{30.86} & \underline{28.66} & \underline{33.88} & \underline{40.80} & \underline{23.64} & \textbf{50.44} \\
UniGraph & 63.21 & 66.73 & 35.88 & 36.28 & 30.92 & 62.61 & 27.04 & 48.57 \\
\ours & \textbf{69.01}$_{\pm 0.41}$ & \textbf{39.11}$_{\pm 0.82}$ & \textbf{35.24}$_{\pm 0.76}$ & \textbf{36.50}$_{\pm 0.84}$ & \textbf{35.77}$_{\pm 1.20}$ & \textbf{70.24}$_{\pm 1.73}$ & \textbf{25.76}$_{\pm 0.84}$ & \underline{47.61}$_{\pm 1.35}$ \\
\bottomrule
\end{tabular}
}
\end{table*}

\textbf{Baselines.}
We compare \ours\ with eight baselines, which can be categorized into four types.
(1) \textbf{MLP} utilizes a multi-layer perceptron to extract deep features individually for each node.
(2) Graph Neural Networks: 
\textbf{GAT}~\cite{GAT} computes the hidden representations of each node in the graph by first learning the importance of its neighbors and then aggregating information from them. 
\textbf{GIN}~\cite{GIN} enhances the representation power of
GNNs by employing a distinct graph encoding method that emphasizes the discrimination of non-isomorphic structures.
(3) Graph Pre-training Models:
\textbf{GraphCL}~\cite{graphcl} utilizes pre-training of graph models through the application of a self-discriminative contrastive learning task on learned node embeddings. 
\textbf{DGI}~\cite{DGI} introduces a self-supervised pre-training task that aims to maximize the mutual information between the local node view and the global graph view.
\textbf{GraphMAE}~\cite{graphmae} present a masked graph autoencoder that mitigates the issues for generative self-supervised graph learning.
(4) graph models:
\textbf{OFA}~\cite{ofa} utilizes a pre-trained LM to align feature spaces of graphs from different domains, enabling supervised training across all graphs.
\textbf{UniGraph} We further include recent graph foundation models such as UniGraph~\cite{unigraph}, which leverages LLMs and GNNs through masked modeling and instruction tuning for generalized graph representations.

\begin{figure*}[h!]
  \centering  \includegraphics[width=0.8\linewidth]{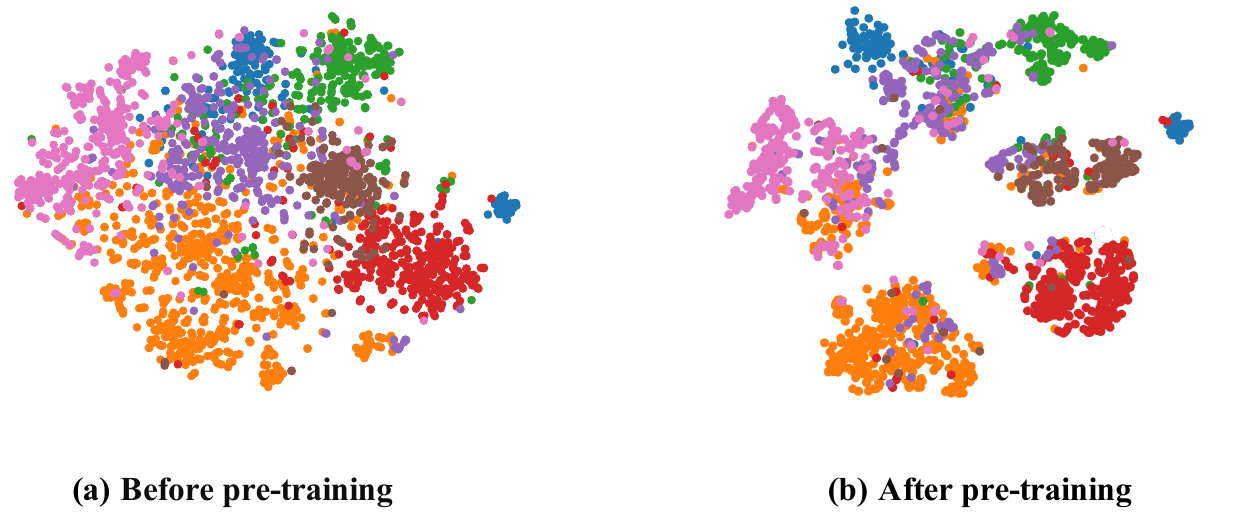}
  \caption{t-SNE visualization on Cora. 
  }
  \label{fg:tsne}
\end{figure*}

\textbf{Experimental Setup.}
We implement \ours\ by PyTorch and conducted the experiments with one A100 GPU. 
The model is optimized by Adam~\cite{kingma2014adam}. 
We employ TAGs for all methods to unify all input data.
Specifically,
we employ the pre-trained language models (LMs) to encode instance features as input for models.
\ours\ and all other self-supervised learning baselines are first pre-trained on the citation network Cora. 
For other self-supervised learning baseline methods,
we train a linear classifier on the top of the embeddings derived from the frozen model.
Then, we evaluate the model performance on the test sets of all seven datasets with different tasks.
We perform a grid search to tune hyper-parameters based on the validation set. 
Details on the search space of these hyper-parameters can be found in Table~\ref{ta:hyper}.

\begin{figure*}[h!]
  \centering  \includegraphics[width=0.9\linewidth]{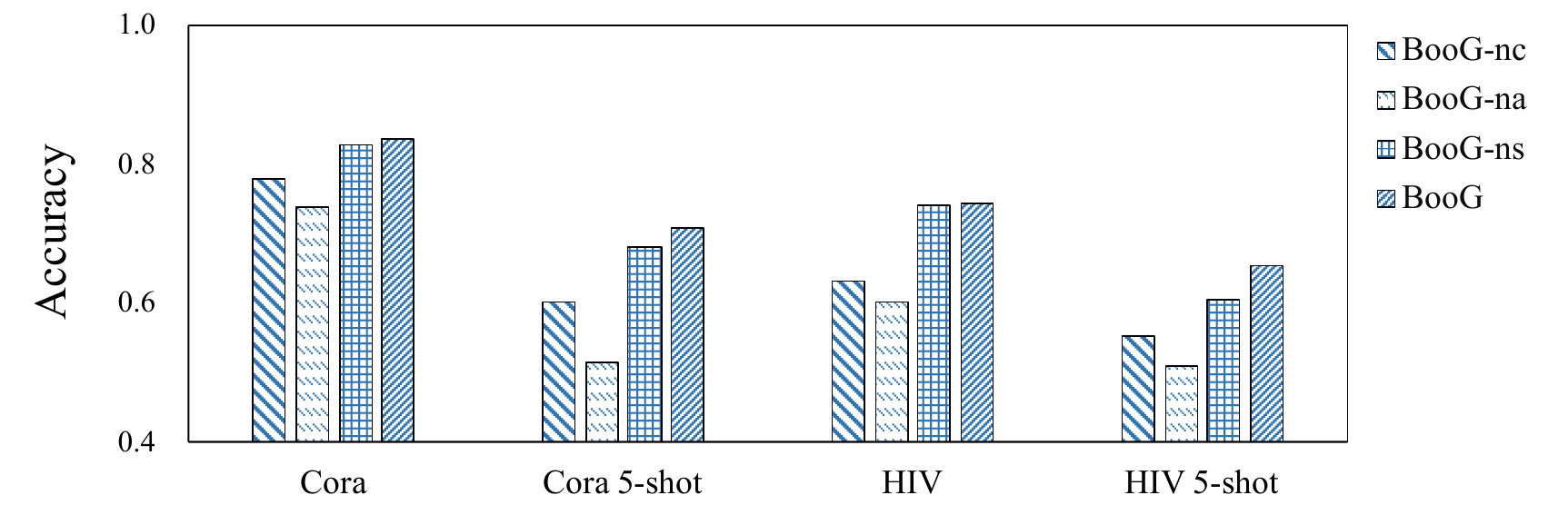}
  \caption{Ablation study.}
  \label{fg:ablation}
\end{figure*}

\subsection{Performance Results}
We next evaluate the model generalizability across various learning scenarios and downstream tasks.
We employ TAGs for all methods to unify all input data.
Specifically,
we employ the pre-trained language models (LMs) to encode instance features as input for models.
\ours\ and all other self-supervised learning baselines are first pre-trained on the citation network Cora.
In the \textbf{supervised learning} setting, 
the parameters of pre-trained models are frozen and the MLP is trained in a supervised manner.
Other baselines are trained directly on the target datasets.
In the \textbf{few-shot learning} setting, 
we freeze the parameters of the pre-trained methods and train their MLPs using a $N$-way $K$-shot task. 
For other methods, 
we train them using the $N$-way $K$-shot task, i.e., they are directly trained on target datasets with labels. 
To accomplish this, we construct a support set by randomly selecting $K$ examples per class from the training split. 
For \textbf{zero-shot learning}, 
different from the above two scenarios, OFA is first pre-trained on the citation network Cora.
For GraphCL, DGI and GraphMAE, 
predictions are generated by comparing the similarity between the instance representations derived from their models and the class label representations encoded by the LM.
Finally, we evaluate all methods on the test sets of the target datasets.

\textbf{Performance analysis.}
Table~\ref{ta:supervise}, Table~\ref{ta:few} and Table~\ref{ta:zero} report the results in supervised learning, few-shot learning, and zero-shot learning, respectively.
From these tables, we observe that:
(1) \ours\ demonstrates the best results in most cases.
(2) The GNN models GAT and GIN perform better in specific domains in supervised learning and few-shot learning. For example, GAT excels on citation network datasets, while GIN is more suited to molecular graphs.
However, these methods perform poorly in other domains because they are typically designed for a single graph domain, especially in few-shot learning.
(3) The three self-supervised learning methods, GraphCL, DGI and GraphMAE exhibit poor domain generalization capabilities in few-shot learning and zero-shot learning,
since they only use the raw graph structure.
(4) OFA demonstrates a certain degree of domain generalization capabilities in few-shot learning, highlighting the effectiveness of in-context learning.
However, \ours\ outperforms OFA in most cases because it constructs unified structural characteristics for graphs from different domains, thereby enhancing the model generalizability.
Overall, 
\ours\ shows strong generalization ability in all scenarios, achieving results comparable to or even better than baseline methods, which proves its effectiveness.

\begin{table*}[h!]
\centering
\normalsize
\caption{Experiment results in supervised learning on different pre-training datasets.
We report accuracy (\%) for node/edge tasks and ROC-AUC score (\%) for graph tasks. 
We highlight the best score on each dataset in bold.
}
\label{other-super}
\resizebox{1\textwidth}{!}{
\begin{tabular}{@{}c|cccccccc@{}}
\toprule
\textbf{Pre-training} & \textbf{Cora} & \textbf{Pubmed} & \textbf{ogbn-arxiv} & \textbf{Wiki-CS} & \textbf{Ele-Fashion} & \textbf{Cora-link} & \textbf{PCBA} & \textbf{HIV} \\ \midrule
Cora & \textbf{83.70}$_{\pm 0.47}$ & 88.51$_{\pm 0.31}$ & 74.57$_{\pm 0.61}$ & 75.84$_{\pm 0.60}$ & 72.34$_{\pm 0.65}$ & \textbf{93.11}$_{\pm 0.41}$ & 58.26$_{\pm 1.27}$ & 74.50$_{\pm 1.42}$  \\
Pubmed & 80.77$_{\pm 0.51}$ & \textbf{89.36}$_{\pm 0.45}$ & 74.89$_{\pm 0.46}$ & 75.07$_{\pm 0.84}$ & 68.02$_{\pm 0.91}$ & 90.65$_{\pm 1.37}$ & 60.25$_{\pm 1.46}$ & 74.11$_{\pm 1.82}$  \\
Wiki-CS & 76.85$_{\pm 0.13}$ & 84.26$_{\pm 0.51}$ & 68.42$_{\pm 0.73}$ & \textbf{79.30}$_{\pm 0.37}$  & 66.74$_{\pm 1.00}$ & 80.16$_{\pm 1.42}$ & 58.86$_{\pm 1.33}$ & \textbf{75.89}$_{\pm 1.64}$\\
Cora$\cup$Pubmed$\cup$Wiki-CS & 83.02$_{\pm 0.58}$ & 88.14$_{\pm 0.45}$ & \textbf{75.36}$_{\pm 0.57}$ & 73.22$_{\pm 0.65}$ & \textbf{74.68}$_{\pm 0.57}$ & 83.26$_{\pm 1.17}$ & \textbf{61.35}$_{\pm 1.14}$ & 75.83$_{\pm 1.77}$  \\
\bottomrule
\end{tabular}
}
\end{table*}

\subsection{Visualization}
We further evaluate the expressiveness of \ours.
In particular,
we
visualize node representations before and after pre-training on Cora with
the t-SNE technique~\cite{van2008visualizing}. 
Colors are employed to differentiate between various node classes, allowing for clear visual distinction. 
Prior to pre-training, the node representations are generated by the language model (LM), which encodes the input data.
As shown in Figure~\ref{fg:tsne}, 
after undergoing pre-training, the node representations corresponding to different classes become more distinctly separated in the representation space. This improvement in separation suggests that the \ours\ model is able to learn richer and more expressive feature representations, capturing more meaningful distinctions between the node classes.

\begin{table*}[h!]
\centering
\normalsize
\caption{Experiment results in few-shot learning on different pre-training datasets.
We report accuracy (\%) for node/edge tasks and ROC-AUC score (\%) for graph tasks. 
We highlight the best score on each dataset in bold.
}
\label{other-few}
\resizebox{1\textwidth}{!}{
\begin{tabular}{@{}cc|ccccccccc@{}}
\toprule
\multicolumn{2}{c|}{\textbf{Pre-training}} & \textbf{Cora} & \textbf{Pubmed} & \textbf{ogbn-arxiv} & \textbf{Wiki-CS} & \textbf{Ele-Fashion} & \textbf{Cora-link} & \textbf{PCBA} & \textbf{HIV} \\ \midrule
\multirow{2}{*}{Cora} & 1-shot & \textbf{44.47}$_{\pm 1.91}$ & 33.85$_{\pm 1.78}$ & 20.18$_{\pm 1.74}$ & 25.11$_{\pm 1.07}$ & \textbf{35.84}$_{\pm 1.21}$ & \textbf{58.27}$_{\pm 1.36}$ & 21.55$_{\pm 1.18}$ & 47.74$_{\pm 1.37}$ \\
 & 5-shot & \textbf{70.82}$_{\pm 1.70}$ & 57.56$_{\pm 1.44}$ & 43.31$_{\pm 1.53}$ & 45.92$_{\pm 0.85}$ & \textbf{54.92}$_{\pm 0.78}$ & \textbf{82.41}$_{\pm 1.65}$ & 28.62$_{\pm 1.54}$ & 65.48$_{\pm 1.02}$ \\ \midrule
\multirow{2}{*}{Pubmed} & 1-shot & 38.91$_{\pm 1.26}$ & \textbf{40.57}$_{\pm 1.31}$ & 24.33$_{\pm 1.52}$ & 24.86$_{\pm 1.44}$ & 32.16$_{\pm 1.10}$ & 54.35$_{\pm 1.45}$ & 23.82$_{\pm 1.45}$ & 48.83$_{\pm 1.44}$ \\
 & 5-shot & 68.25$_{\pm 1.04}$ & \textbf{65.86}$_{\pm 1.27}$ & 45.16$_{\pm 1.31}$ & 46.03$_{\pm 1.15}$ & 49.03$_{\pm 0.92}$ & 79.30$_{\pm 1.58}$ & 35.41$_{\pm 1.61}$ & 65.80$_{\pm 1.32}$ \\ \midrule
\multirow{2}{*}{Wiki-CS} & 1-shot & 31.42$_{\pm 1.34}$ & 28.46$_{\pm 1.21}$ & 21.65$_{\pm 1.82}$ & \textbf{31.62}$_{\pm 1.03}$ & 29.70$_{\pm 1.05}$ & 49.33$_{\pm 1.42}$ & 23.67$_{\pm 1.48}$ & 47.16$_{\pm 1.35}$ \\
 & 5-shot & 54.28$_{\pm 1.77}$ & 53.81$_{\pm 1.45}$ & 43.88$_{\pm 1.64}$ & \textbf{62.16}$_{\pm 1.15}$ & 50.80$_{\pm 0.85}$ & 60.36$_{\pm 1.85}$ & 36.12$_{\pm 1.53}$ & 64.93$_{\pm 1.48}$ \\ \midrule
\multirow{2}{*}{Cora$\cup$Pubmed$\cup$Wiki-CS} & 1-shot & 37.22$_{\pm 1.03}$ & 31.74$_{\pm 1.18}$ & \textbf{33.60}$_{\pm 1.47}$ & 30.41$_{\pm 1.82}$ & 33.10$_{\pm 1.13}$ & 52.87$_{\pm 1.71}$ & \textbf{25.82}$_{\pm 1.46}$ & \textbf{49.26}$_{\pm 1.16}$ \\
 & 5-shot & 60.16$_{\pm 1.84}$ & 58.32$_{\pm 1.41}$ & \textbf{48.26}$_{\pm 1.15}$ & 60.85$_{\pm 1.37}$ & 52.45$_{\pm 0.94}$ & 80.30$_{\pm 1.14}$ & \textbf{39.36}$_{\pm 1.28}$ & \textbf{68.30}$_{\pm 1.24}$ \\
\bottomrule
\end{tabular}
}
\end{table*}

\begin{table*}[h!]
\centering
\normalsize
\caption{Experiment results in zero-shot learning on different pre-training datasets.
We report accuracy (\%) for node/edge tasks and ROC-AUC score (\%) for graph tasks. 
We highlight the best score on each dataset in bold.
}
\label{other-zero}
\resizebox{1\textwidth}{!}{
\begin{tabular}{@{}c|cccccccc@{}}
\toprule
\textbf{Pre-training} & \textbf{Cora} & \textbf{Pubmed} & \textbf{ogbn-arxiv} & \textbf{Wiki-CS} & \textbf{Ele-Fashion} & \textbf{Cora-link} & \textbf{PCBA} & \textbf{HIV} \\ \midrule
Cora & \textbf{69.01}$_{\pm 0.41}$ & 39.11$_{\pm 0.82}$ & 35.24$_{\pm 0.76}$ & 36.50$_{\pm 0.84}$ & 35.77 $_{\pm 1.20}$ & \textbf{70.24}$_{\pm 1.73}$ & 25.76$_{\pm 0.84}$ & 47.61$_{\pm 1.35}$ \\
Pubmed & 63.21$_{\pm 1.14}$ & 66.73$_{\pm 1.27}$ & 35.88$_{\pm 1.43}$ & 36.28$_{\pm 0.79}$ & 30.92$_{\pm 0.93}$ & 62.61$_{\pm 1.85}$ & 27.04$_{\pm 0.78}$ & 48.57$_{\pm 1.65}$ \\
Wiki-CS & 43.81$_{\pm 0.82}$ & 45.64$_{\pm 1.12}$ & 30.17$_{\pm 1.52}$ & \textbf{57.32}$_{\pm 0.67}$ & 29.42$_{\pm 0.86}$ & 60.21$_{\pm 1.57}$ & 27.13$_{\pm 0.85}$ & 47.62$_{\pm 1.37}$ \\
Cora$\cup$Pubmed$\cup$Wiki-CS & 65.71$_{\pm 1.37}$ & \textbf{67.35}$_{\pm 0.92}$ & \textbf{38.24}$_{\pm 1.33}$ & 55.43$_{\pm 0.81}$ & \textbf{36.88}$_{\pm 1.02}$ & 65.71$_{\pm 1.24}$ & \textbf{31.82}$_{\pm 0.54}$ & \textbf{53.71}$_{\pm 1.25}$ \\
\bottomrule
\end{tabular}
}
\end{table*}

\subsection{Ablation Study}
\label{ap:ablation}
The ablation study is performed to understand the importance of main components of \ours~in supervised learning and 5-shot learning scenarios.
We select Cora and HIV as two representative datasets.
The model is pre-trained on Cora. 
We first remove the class labels from the super nodes and call this variant \ours-nc (\textbf{n}o \textbf{c}lass label). 
Accordingly, 
in the pre-training task, 
we use the anchor node before aggregation as the positive sample and treat other instances of dataset as negative samples.
Secondly,
we remove the anchor nodes from the super nodes and treat the anchor node as a member of the contextual neighborhood.
We call this variant \ours-na (\textbf{n}o \textbf{a}nchor node). 
Further, 
we remove the similarity matching process and concatenate the representations based on different class labels as the final representation of the input to the MLP.
We call this variant \ours-ns (\textbf{n}o \textbf{s}imilarity matching).
We compare \ours\ with these three variants, and the results are presented in Figure~\ref{fg:ablation}.
Our findings show that \ours\ outperforms all the variants on the two datasets.
Further, the decline in performance exhibited by both \ours-nc and \ours-na highlights the importance of super nodes.
In other words,
the super nodes significantly enhance model generalizability, especially in graph domain transfer,
further demonstrated the effectiveness of \ours.

\subsection{Performance on different pre-training datasets}
We next evaluate the model generalizability across various
learning scenarios and downstream tasks based on different pre-training datasets. 
\ours\ is first pre-trained on Cora/Pubmed/Wiki-CS and their union.
In the \textbf{supervised learning} setting, 
the parameters of pre-trained models are frozen and the MLP is trained in a supervised manner.
In the \textbf{few-shot learning} setting, 
we freeze the parameters of the pre-trained methods and train their MLPs using a $N$-way $K$-shot task. 
We evaluate \ours\ on the test sets of the target datasets.

Table~\ref{other-super}, Table~\ref{other-few} and Table~\ref{other-zero} report the results in supervised learning, few-shot learning,
and zero-shot learning, respectively.
From these tables, we observe that:
(1) \ours\ achieves good performance on various pre-training datasets.
(2) Compared to other target datasets, \ours\ demonstrates better generalization on the pre-training source dataset. 
(3) Joint pre-training on multiple datasets performs better on unseen-domain datasets compared to pre-training on a single dataset.
Overall, \ours\ shows strong generalization
ability in all scenarios on different pre-training datasets,  which proves its effectiveness.

\begin{figure*}[h!]
  \centering  \includegraphics[width=0.8\linewidth]{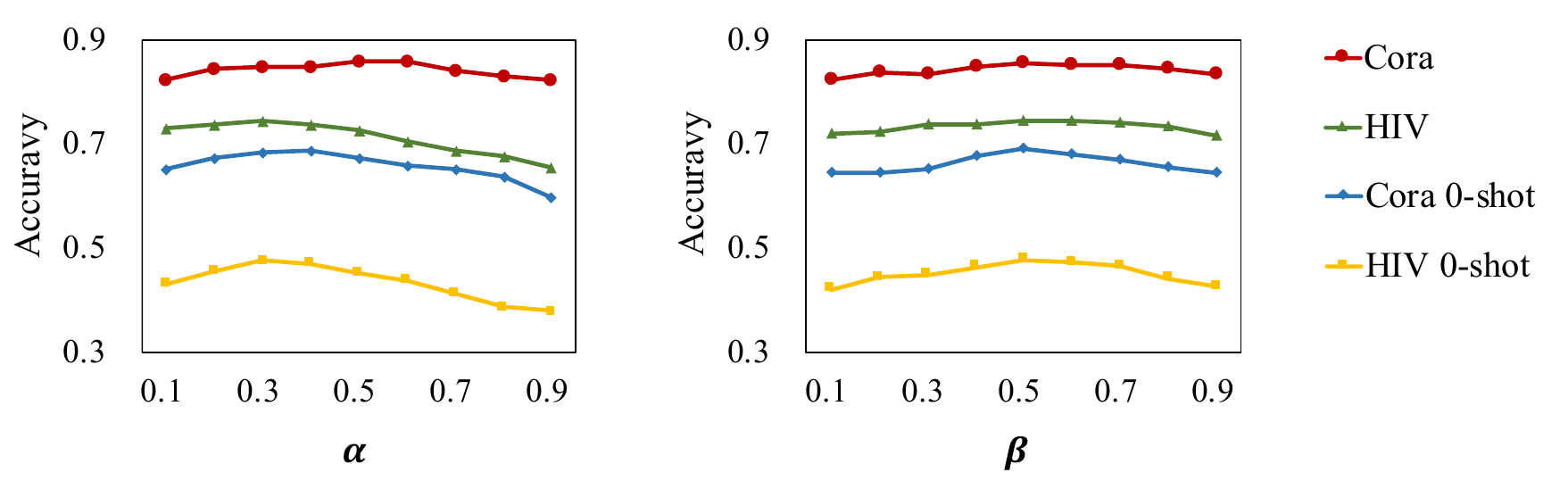}
  \caption{Hyper-parameter sensitivity analysis.
  }
  \label{fg:hyper}
\end{figure*}

\begin{table*}[t]
\centering
\caption{Comparison of GPU hours and performance on Cora. All time measured on a single A100 GPU.
}
\label{tab:cora_efficiency}
\resizebox{0.75\textwidth}{!}{
\begin{tabular}{cccc}
\toprule
\textbf{Method} & \textbf{Pre-training} & \textbf{Downstream Training} & \textbf{Inference}  \\
\midrule

GAT & - & 23 min (supervised) / 5 min (5-shot) & 1 min  \\
GraphCL & 52 min & 18 min (supervised) / 4 min (5-shot) & 1 min \\
GraphMAE & 63 min & 25 min (supervised) / 5 min (5-shot) & 1 min  \\
UniGraph & 2.8 h & 8 min (supervised) / 5 min (5-shot) & 2 min  \\
BooG & 43 min & 5 min (supervised) / 2 min (5-shot) & 1 min  \\
\bottomrule
\end{tabular}
}
\end{table*}

\subsection{Sensitivity Analysis}
\label{ap:hyper-analysis}
In this section, we analyze the sensitivity of BooG to the hyper-parameters $\alpha$ and $\beta$, which controls the influence of the class label embedding in constructing super nodes. Specifically, $\alpha$ determines the balance between the anchor node’s original representation and the class label information when forming the super node embedding. A larger $\alpha$ places more emphasis on the class label prior, while a smaller $\alpha$ makes the super node more similar to the original anchor node embedding.
As shown in Figure~\ref{fg:hyper}, the performance of BooG exhibits sensitivity to the choice of $\alpha$. Moderate values of $\alpha$ typically yield better results, as they allow the model to effectively balance the anchor node’s instance-specific semantics and the global class-level prior. If $\alpha$ is too small, the super node degenerates to the original node representation, reducing the benefit of class-conditioned modeling. Conversely, if $\alpha$ is too large, the model overemphasizes class priors and may ignore useful instance-specific signals. Unlike vanilla GNNs, BooG explicitly leverages class priors via super nodes, and $\alpha$ controls this mechanism.
For hyper-parameter $\beta$, \ours\ consistently performs well across a wide range of values, which demonstrates that $\beta$ does not significantly impact the performance of \ours.

\subsection{Efficiency Analysis}
To further address practical efficiency, we compare the GPU time required for both pre-training and downstream tasks on the Cora dataset. Table~\ref{tab:cora_efficiency} reports the results.
We observe that \ours\ requires significantly less pre-training time (only 43 minutes) compared to existing graph pre-training methods such as GraphCL (52 minutes), GraphMAE (63 minutes), and UniGraph (2.8 hours). This efficiency is attributed to BooG's lightweight design, which avoids costly full-graph augmentation or reconstruction processes and instead leverages text semantics and localized graph neighborhoods.

For downstream tasks, BooG also achieves fast tuning and inference. The MLP-based classifier on BooG’s representations requires only 5 minutes for supervised tuning, and even faster under 5-shot settings (2 minute). The inference process relies on simple similarity matching, making BooG extremely efficient at deployment.
These results highlight that BooG provides a favorable balance between efficiency and generalization.

\section{Conclusion}
In this paper, 
we boost graph models from structural perspective and propose \ours. 
BooG boosts model’s cross-domain and cross-task generalization ability from two key aspects. 
First, BooG unifies graph attributes by encoding node texts of TAGs and class label descriptions into a shared space using a pre-trained LM, thereby
equipping the model to handle graph data from diverse domains.
More critically, BooG unifies graph structures by introducing the concept of sub-graphs and super nodes. Sub-graphs offer a consistent task template across all levels of tasks, consisting of anchor nodes and their neighbor nodes. 
The super nodes,
along with virtual edges, establish a standardized aggregation mechanism that fuses rich information from neighborhoods and associated class labels, accommodating graph structural characteristics inherent to different domains.
These designs integrate a supernode-centered graph structure that encapsulates label preferences, making \ours\ compatible with different graph domains and task levels.
Our extensive experiments on seven datasets, covering a range of domains and tasks, show that BooG outperforms state-of-the-art competitors in most cases.

\begin{acknowledgement}
This work is supported by National Natural Science Foundation
of China No. 62202172.
\end{acknowledgement}




\bibliographystyle{fcs}
\bibliography{ref}


\begin{biography}
{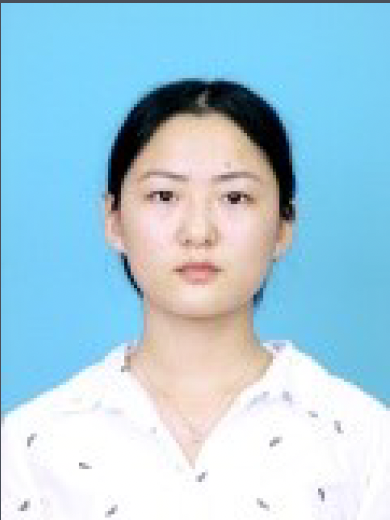}{Yao~Cheng} received the BS degree from Hainan University, Hainan, China. She is currently working toward the PhD degree with East China Normal University, Shanghai, China. With a focus on Artificial Intelligence, Graph Neural Network, and Deep Learning. 

\end{biography}

\vspace{15mm}

\begin{biography}
{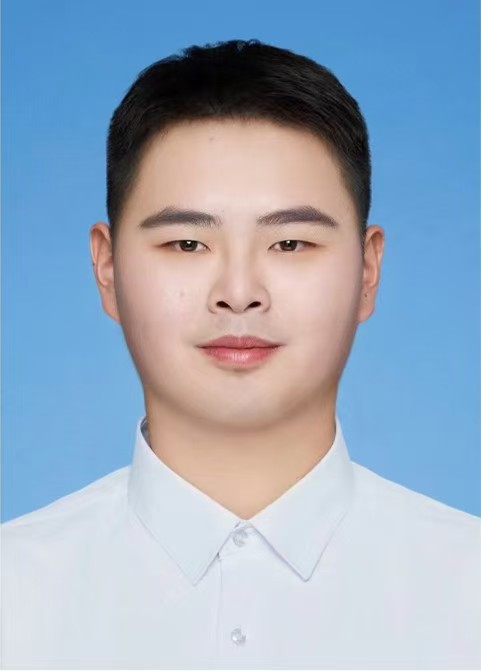}{Jiapeng~Zhu} is a first-year Ph.D. student in the combined master's and doctoral program at the School of Data Science and Engineering, East China Normal University. He received his B.S. degree in Software Engineering from East China University of Science and Technology in 2022. His research interests include Deep Reinforcement Learning, Graph Representation Learning, and Large Language Models.
\end{biography}

\begin{biography}
{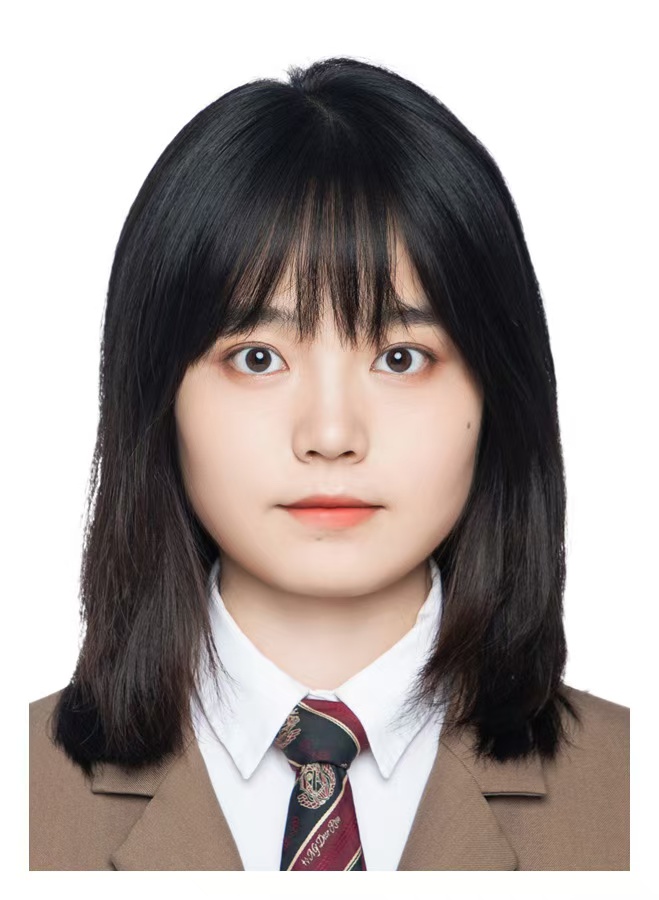}{Yige~Zhao} earned her Master’s degree from the School of Data Science and Engineering at East China Normal University. Her research interests include graph neural networks and large language models.
\end{biography}

\begin{biography}
{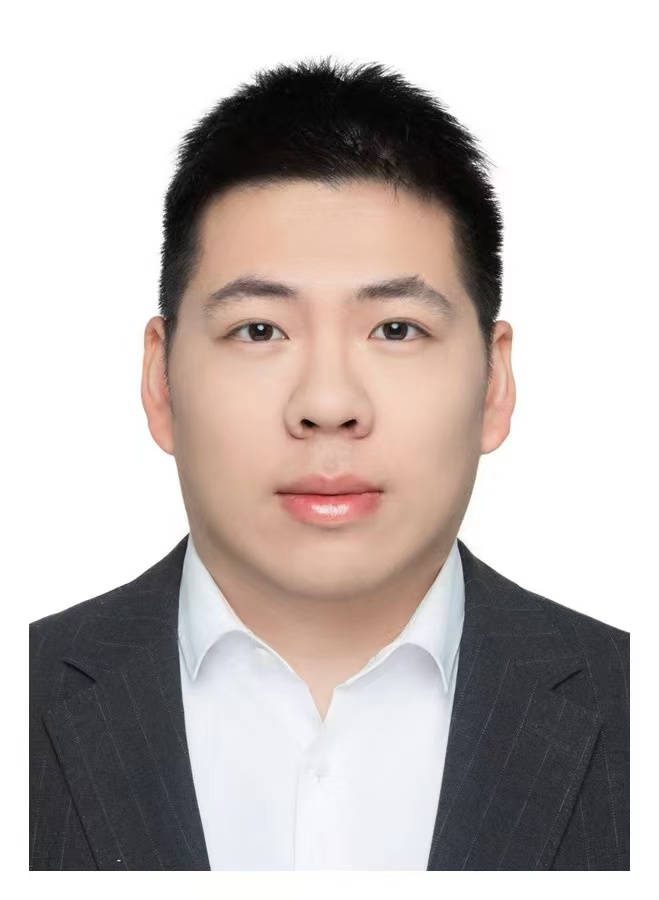}{Jianxiang~Yu}
is currently a PhD student at the School of Data Science and Engineering in East China Normal University.
His research interests include graph neural networks, data mining and large language models.
\end{biography}

\vspace{15mm}

\begin{biography}
{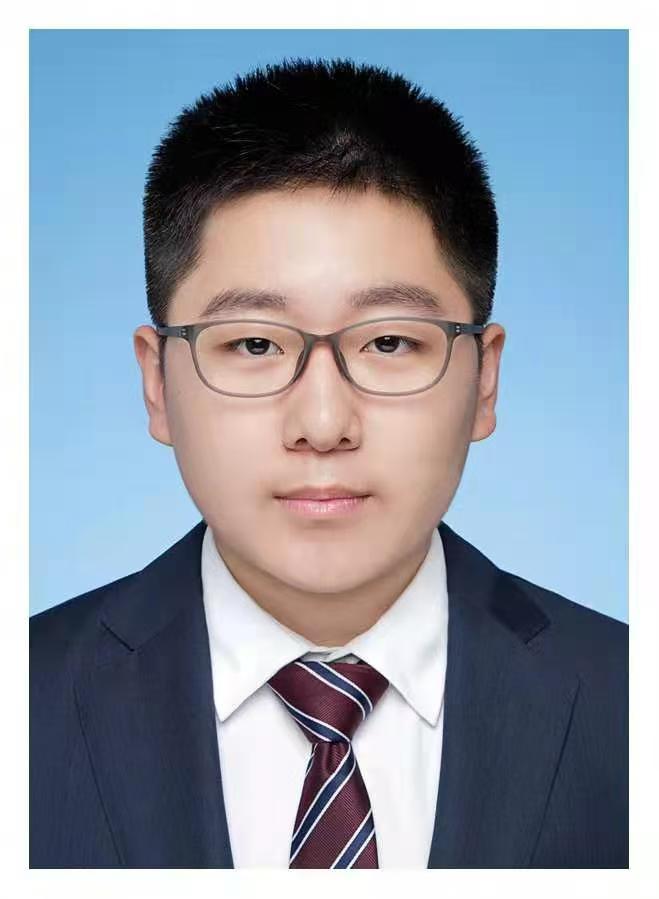}{Jiaqi~Tan} is currently working toward the MS degree at the School of Data Science and Engineering in East China Normal University. His general research interests include graph neural networks and large language models.
\end{biography}


\begin{biography}
{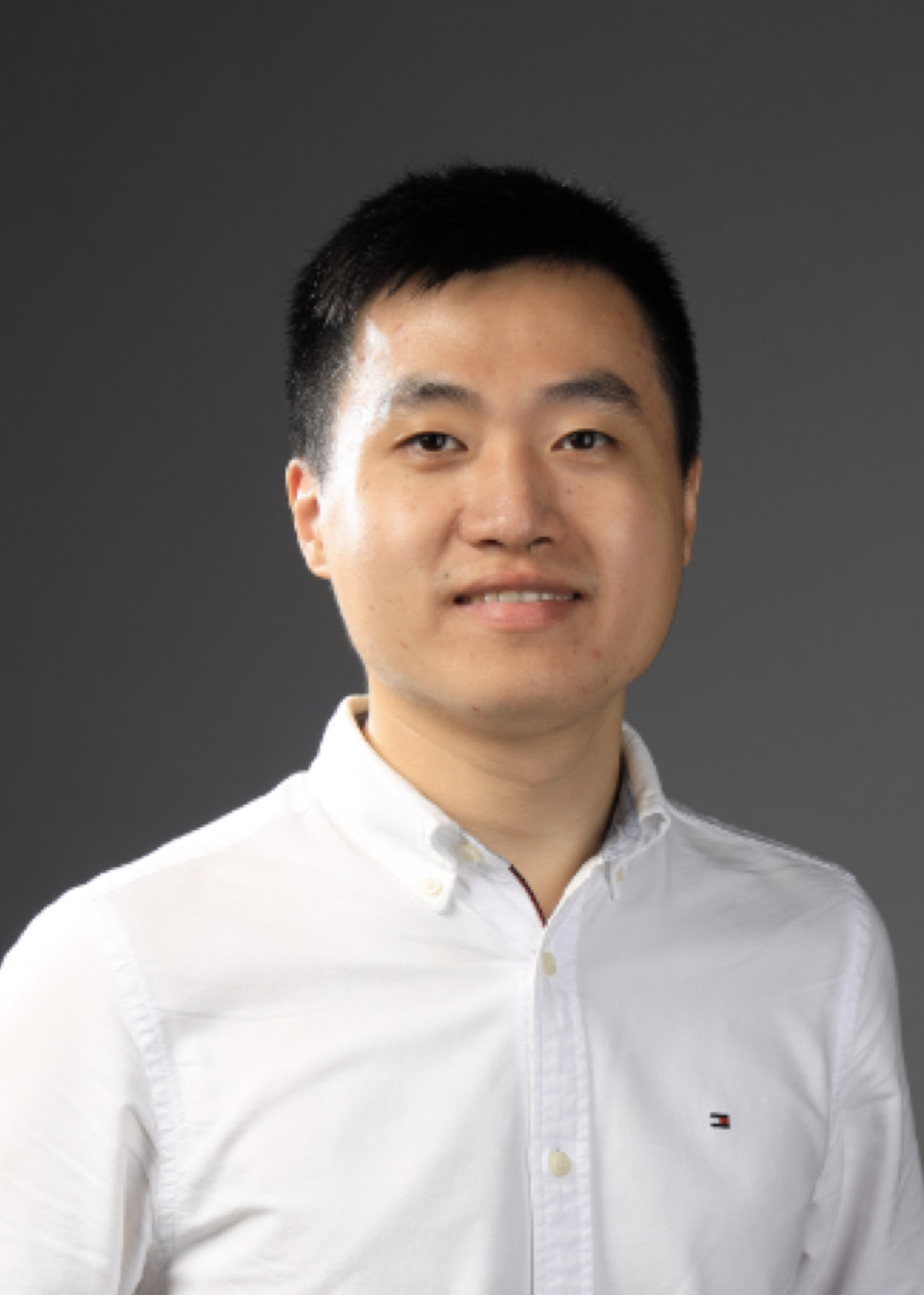}{Xiang~Li} received his Ph.D. degree from the University of Hong Kong in 2018. From 2018 to 2020, 
he worked as a research scientist in the Data Science Lab at JD.com and a research associate at The University of Hong Kong, respectively. 
He is currently a research professor at the School of Data Science and Engineering in East China Normal University. 
His general research interests include data mining and machine learning applications.
\end{biography}

\end{document}